%% file: main.tex
\documentclass[conference]{IEEEtran}
\IEEEoverridecommandlockouts
\usepackage{cite}
\usepackage{amsmath,amssymb,amsfonts}
\usepackage{algorithm}
\usepackage{algorithmic}
\usepackage{graphicx}
\usepackage{textcomp}
\usepackage{xcolor}
\usepackage{booktabs}
\usepackage{hyperref}
\usepackage{listings}

\definecolor{codegreen}{rgb}{0.25,0.5,0.35}
\definecolor{codepurple}{rgb}{0.58,0,0.82}
\definecolor{codeblue}{rgb}{0.13,0.29,0.53}
\definecolor{codegray}{rgb}{0.5,0.5,0.5}
\definecolor{backcolor}{rgb}{0.97,0.97,0.97}

\def\BibTeX{{\rm B\kern-.05em{\sc i\kern-.025em b}\kern-.08em
    T\kern-.1667em\lower.7ex\hbox{E}\kern-.125emX}}

\newcommand{\bheading}[1]{{\noindent{\textbf{#1}}}}

\begin{document}

\title{Code Evolution for Control: Synthesizing Policies via LLM-Driven Evolutionary Search}

\author{
    \IEEEauthorblockN{
        Ping Guo\textsuperscript{1,2},
        Chao Li\textsuperscript{1},
        Yinglan Feng\textsuperscript{3},
        Chaoning Zhang\textsuperscript{4}
    }
    \IEEEauthorblockA{
        \textsuperscript{1}Shenzhen Yuyi Tech. Co. Ltd., Shenzhen, China\\
        \textsuperscript{2}City University of Hong Kong, Hong Kong\\
        \textsuperscript{3}Shenzhen University, Shenzhen, China\\
        \textsuperscript{4}University of Electronic Science and Technology of China, Chengdu, China\\
        pguo6680@gmail.com, 1005044057@qq.com, yinglfeng@szu.edu.cn, chaoningzhang1990@gmail.com
    }
}

\maketitle

\input{secs/00-abstract.tex}


\input{secs/01-introduction.tex}
\input{secs/02-related.tex}
\input{secs/03-methodology.tex}
\input{secs/04-experiments.tex}
\input{secs/05-results.tex}
\input{secs/06-discussion.tex}

\bibliographystyle{IEEEtran}
\bibliography{main}

\end{document}

%% file: secs/00-abstract.tex
\begin{abstract}
    Designing effective control policies for autonomous systems remains a fundamental challenge, traditionally addressed through reinforcement learning or manual engineering.
    While reinforcement learning has achieved remarkable success, it often suffers from high sample complexity, reward shaping difficulties, and produces opaque neural network policies that are hard to interpret or verify.
    Manual design, on the other hand, requires substantial domain expertise and struggles to scale across diverse tasks.
    In this work, we demonstrate that LLM-driven evolutionary search can effectively synthesize interpretable control policies in the form of executable code.
    By treating policy synthesis as a code evolution problem, we harness the LLM's prior knowledge of programming patterns and control heuristics while employing evolutionary search to explore the solution space systematically.
    We implement our approach using EvoToolkit, a framework that seamlessly integrates LLM-driven evolution with customizable fitness evaluation. Our method iteratively evolves populations of candidate policy programs, evaluating them against task-specific objectives and selecting superior individuals for reproduction. This process yields compact, human-readable control policies that can be directly inspected, modified, and formally verified.
    Experimental results on the LunarLander benchmark demonstrate that our approach achieves a 70\% landing success rate, surpassing the 60\% of PPO, while producing compact 59-line policies that can be directly inspected and verified. Although the average reward is lower than neural network policies, the evolved code exhibits superior interpretability and reliability in safety-critical metrics. This work highlights the potential of combining foundation models with evolutionary computation for synthesizing trustworthy control policies in autonomous systems. Code is available at \url{https://github.com/pgg3/EvoControl}.
\end{abstract}

%% file: secs/01-introduction.tex
\section{Introduction}

Designing effective control policies for autonomous systems remains a fundamental challenge in robotics and artificial intelligence. Traditional approaches fall into two categories: reinforcement learning (RL) and manual engineering. While RL methods like Proximal Policy Optimization (PPO)~\cite{schulman:2017:ppo} have achieved remarkable success, they suffer from high sample complexity, reward shaping difficulties, and produce opaque neural network policies that resist interpretation and verification~\cite{duan:2016:rl}. Manual design, conversely, requires substantial domain expertise and struggles to scale across diverse tasks.

For safety-critical applications, the inherent opacity of neural network policies presents a fundamental obstacle. When failures occur, diagnosing the root cause is often intractable.
This precludes formal verification, complicates systematic debugging, and ultimately undermines trust in deployed autonomous systems.

To address these limitations of neural network policies, recent research has explored \emph{programmatic policies}, which express control logic as executable code~\cite{verma:2018:programmatic,trivedi:2021:programmatic}.
Such policies provide inherent transparency: they can be directly inspected, modified, and formally verified.
However, synthesizing effective programmatic policies remains difficult.
Manual implementation demands significant domain expertise, while automated approaches face their own obstacles: designing domain-specific languages requires substantial upfront effort, and traditional program search methods like genetic programming are prone to combinatorial explosion~\cite{kronberger2024inefficiency}.

LLM-driven code evolution has recently achieved notable success across diverse domains.
FunSearch~\cite{romera:2024:funsearch} discovered novel mathematical constructions by evolving program components, while CoEvo~\cite{guo:2024:coevo} extended this paradigm to symbolic regression with a dynamic knowledge library for open-ended discovery.
In the optimization domain, Evolution of Heuristics~\cite{liu:2024:eoh} applied LLM-based mutation and crossover to automatically design algorithms, and EvoEngineer~\cite{guo:2025:evoengineer} optimized CUDA kernels through reflection-guided code evolution.
These methods demonstrate that LLMs can serve as intelligent search operators, generating semantically meaningful variations that traditional approaches struggle to produce.
However, their application to control policy synthesis remains unexplored.

This observation motivates the following research questions:
\begin{itemize}
    \item \textbf{RQ1:} Can LLM-driven evolution synthesize control policies that match or exceed RL baselines?
    \item \textbf{RQ2:} How do different LLM evolution strategies compare for policy synthesis?
    \item \textbf{RQ3:} Do evolved code policies offer interpretability advantages over neural networks?
\end{itemize}

To address these questions, we propose treating policy synthesis as a code evolution problem.
The LLM serves dual roles: as a mutation operator that generates semantically meaningful code variations, and as a crossover operator that recombines promising components across individuals.
This approach harnesses the LLM's prior knowledge of programming patterns and control heuristics while using evolutionary search to systematically explore the solution space.

Our contributions are as follows:
\begin{itemize}
    \item We formalize control policy synthesis as a code evolution problem and demonstrate its feasibility using LLM-driven evolutionary search.
    \item We systematically compare three LLM evolution strategies (FunSearch, EoH, and EvoEngineer) on the LunarLander control benchmark.
    \item We show that evolved policies achieve competitive performance with significantly better interpretability than neural network alternatives.
\end{itemize}

%% file: secs/02-related.tex
\section{Related Work}

\subsection{Reinforcement Learning for Control}

Reinforcement learning has become the dominant paradigm for learning control policies in robotics and autonomous systems.
Value-based methods like DQN~\cite{mnih:2015:dqn} learn action-value functions,
while policy gradient methods like PPO~\cite{schulman:2017:ppo} directly optimize parameterized policies.
Despite impressive results, RL methods typically require millions of environment interactions
and produce opaque neural network policies that resist interpretation~\cite{duan:2016:rl}.

\subsection{Program Synthesis and Neuro-Symbolic Methods}

Program synthesis aims to automatically generate programs from specifications~\cite{gulwani:2017:program}. Recent neuro-symbolic approaches combine neural networks with symbolic reasoning to improve interpretability~\cite{garcez:2023:neurosymbolic}. However, these methods often require domain-specific languages or extensive supervision.

\subsection{LLM-Driven Code Evolution}

Large language models have demonstrated remarkable code generation capabilities~\cite{chen:2021:codex,li:2023:starcoder,roziere:2023:codellama}, and recent work has begun exploring their use to iteratively improve code~\cite{yang:2024:opro,liu2024systematic}.

A particularly promising direction combines LLMs with evolutionary computation. FunSearch~\cite{romera:2024:funsearch} pioneered this approach, using LLMs to evolve programs that discovered new mathematical constructions with improved bounds for the cap set problem. CoEvo~\cite{guo:2024:coevo} extended this paradigm to symbolic regression, introducing a dynamic knowledge library that enables continual, open-ended discovery. Evolution of Heuristics (EoH)~\cite{liu:2024:eoh} employed LLMs as mutation and crossover operators for combinatorial optimization, demonstrating that semantic code transformations significantly outperform random perturbations. EvoEngineer~\cite{guo:2025:evoengineer} applied reflection-guided LLM evolution to CUDA kernel optimization, achieving substantial speedups over baseline implementations.

Our work extends this line of research to control policy synthesis, where the goal is to evolve interpretable Python functions that map states to actions.

%% file: secs/03-methodology.tex
\section{Methodology}

\subsection{Problem Formulation}

We formalize control policy synthesis as a code evolution problem. Given an environment $\mathcal{E}$ with state space $\mathcal{S}$, action space $\mathcal{A}$, and reward function $R: \mathcal{S} \times \mathcal{A} \rightarrow \mathbb{R}$, we seek to find a policy function $\pi: \mathcal{S} \rightarrow \mathcal{A}$ that maximizes the expected cumulative reward:

\begin{equation}
    \pi^* = \arg\max_{\pi} \mathbb{E}\left[\sum_{t=0}^{T} R(s_t, \pi(s_t))\right]
\end{equation}

Unlike neural network policies, we constrain $\pi$ to be an interpretable Python function that can be directly inspected, modified, and verified.

\subsection{LLM-Driven Evolution Framework}

Our approach maintains a population $P = \{\pi_1, \pi_2, \ldots, \pi_n\}$ of candidate policy programs and evolves them using LLM-based operators. The overall procedure is summarized in Algorithm~\ref{alg:evolution}.

\begin{algorithm}[t]
    \caption{LLM-Driven Policy Evolution}
    \label{alg:evolution}
    \begin{algorithmic}[1]
        \REQUIRE Environment $\mathcal{E}$, population size $n$, generations $G$
        \ENSURE Best policy $\pi^*$
        \STATE Initialize population $P \leftarrow \{\pi_1, \ldots, \pi_n\}$ via LLM
        \FOR{$g = 1$ to $G$}
        \FOR{each $\pi_i \in P$}
        \STATE $f(\pi_i) \leftarrow$ \textsc{Evaluate}($\pi_i$, $\mathcal{E}$)
        \ENDFOR
        \STATE $P_{\text{parents}} \leftarrow$ \textsc{Select}($P$, $f$)
        \STATE $P_{\text{offspring}} \leftarrow$ \textsc{LLM-Evolve}($P_{\text{parents}}$)
        \STATE $P \leftarrow$ \textsc{Replace}($P$, $P_{\text{offspring}}$, $f$)
        \ENDFOR
        \RETURN $\arg\max_{\pi \in P} f(\pi)$
    \end{algorithmic}
\end{algorithm}

The key innovation lies in \textsc{LLM-Evolve} (line 7), which uses LLMs to generate semantically meaningful code transformations. We compare three strategies for this operator:

\bheading{FunSearch}~\cite{romera:2024:funsearch} uses a single-prompt approach, showing the LLM a progression from worse to better solutions and asking it to continue the improvement trajectory.

\bheading{EoH}~\cite{liu:2024:eoh} employs five specialized operators (initialization, exploration, guided crossover, structural mutation, and parameter mutation), enabling both exploration and exploitation of the solution space.

\bheading{EvoEngineer}~\cite{guo:2025:evoengineer} provides rich context including episode statistics, failure modes, and domain-specific hints, requiring structured responses with code and rationale.
\subsection{Fitness Evaluation}

For control tasks, fitness is computed as the average reward over multiple episodes:

\begin{equation}
    f(\pi) = \frac{1}{K}\sum_{k=1}^{K} \sum_{t=0}^{T_k} R(s_t^{(k)}, \pi(s_t^{(k)}))
\end{equation}

where $K$ is the number of evaluation episodes and $T_k$ is the episode length.

Policies are executed in a sandboxed environment with restricted builtins to ensure safety. Invalid actions or runtime errors result in minimum fitness scores.

%% file: secs/04-experiments.tex
\section{Experiments}

\subsection{Experimental Setup}

\bheading{Task Environment.}
We evaluate on LunarLander-v3 from Gymnasium~\cite{towers:2024:gymnasium}, a classical control benchmark where the agent must land a spacecraft on a designated pad. The environment provides an 8-dimensional continuous state (position, velocity, angle, angular velocity, leg contacts) and 4 discrete actions (do nothing, fire left/main/right engine). The reward function incentivizes precise landing (+100 to +140) while penalizing crashes (-100) and fuel consumption. An episode achieving cumulative reward $\geq 200$ is considered successful.

\bheading{Baselines.}
We compare three LLM-driven evolution strategies against two baselines:
(1) \textit{Random}, which selects actions uniformly at random;
(2) \textit{PPO}~\cite{schulman:2017:ppo}, trained for 1M timesteps using Stable-Baselines3~\cite{raffin:2021:sb3} with default hyperparameters.
The three LLM-based methods (FunSearch, EoH, and EvoEngineer) are configured identically for fair comparison.

\bheading{Configuration.}
All LLM evolution experiments use \texttt{GPT-4o} as the base model with temperature 0.7. We maintain a population of 10 candidate policies and evolve for 20 generations. Each policy's fitness is computed as the average reward over 10 evaluation episodes. Results are averaged over 5 independent runs with different random seeds.

\bheading{Metrics.}
We assess methods along three dimensions:
(1) \textit{Performance}: average episode reward and success rate (reward $> 200$);
(2) \textit{Efficiency}: number of LLM API calls and environment episodes consumed;
(3) \textit{Interpretability}: lines of code and cyclomatic complexity of the final policy.

%% file: secs/05-results.tex
\section{Results}

\subsection{Main Results}

Table~\ref{tab:main_results} summarizes the performance of all methods on LunarLander-v3.

\begin{table}[t]
    \centering
    \caption{Performance comparison on LunarLander-v3. Results for LLM methods are averaged over 5 seeds.}
    \label{tab:main_results}
    \begin{tabular}{lcccc}
        \hline
        \textbf{Method} & \textbf{Avg Reward} & \textbf{Success\%} & \textbf{LLM Calls} & \textbf{LoC} \\
        \hline
        Random          & $-200 \pm 50$       & 0\%                & -                  & N/A          \\
        PPO (1M steps)  & $214 \pm 27$        & 60\%               & -                  & N/A          \\
        \hline
        FunSearch       & $9.1 \pm 33.8$      & 22\%               & $\sim$45           & 20-35        \\
        EoH             & $6.2 \pm 51.2$      & 16\%               & $\sim$45           & 25-40        \\
        EvoEngineer     & $66.6 \pm 56.1$     & 40\%               & $\sim$45           & 30-50        \\
        \hline
        EvoEngineer+    & $\mathbf{143.6}$    & $\mathbf{70\%}$    & $\sim$200          & 59           \\
        \hline
    \end{tabular}
\end{table}

\bheading{Comparison among LLM-based methods.}
Among the three evolution strategies, EvoEngineer significantly outperforms both FunSearch and EoH. With an identical budget of $\sim$45 LLM calls, EvoEngineer achieves an average reward of 66.6 and 40\% success rate, while FunSearch and EoH remain near zero (9.1 and 6.2 respectively). This 7$\times$ improvement in average reward demonstrates that providing rich context (episode statistics, failure modes, domain hints) is crucial for effective policy evolution. The structured prompts in EvoEngineer enable the LLM to make more informed code modifications, whereas simpler approaches struggle to discover effective control strategies.

\bheading{Comparison with PPO.}
When extended to 200 LLM calls (EvoEngineer+), consuming approximately 1M environment steps comparable to PPO's training budget, our approach achieves 143.6 average reward. More importantly, EvoEngineer+ attains a \textbf{70\% success rate, exceeding PPO's 60\%}. While PPO achieves higher average reward due to more consistent performance across episodes, the evolved policy demonstrates superior landing success.

\bheading{The interpretability advantage.}
The key distinction is that EvoEngineer+ produces a 59-line Python function that can be directly inspected, understood, and modified by human engineers. In contrast, PPO's policy is encoded in $\sim$10K neural network parameters that offer no insight into the decision-making logic. This interpretability comes at modest performance cost (67\% of PPO's average reward) while achieving higher task success rate, representing an attractive trade-off for safety-critical applications where understanding policy behavior is essential.

\subsection{Convergence Analysis}

\begin{figure}[t]
    \centering
    \includegraphics[width=0.96\columnwidth]{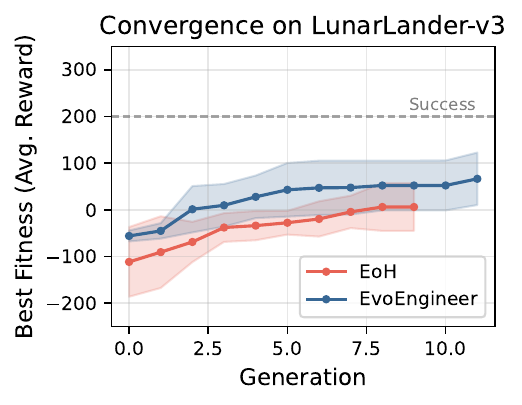}
    \caption{Best fitness vs. generation for EoH and EvoEngineer. Shaded regions indicate standard deviation across 5 seeds.}
    \label{fig:convergence}
\end{figure}

Figure~\ref{fig:convergence} shows the convergence curves of EoH and EvoEngineer.\footnote{EoH runs for 10 generations while EvoEngineer runs for 12, as EoH uses 5 operators per generation versus EvoEngineer's 3, resulting in different generation counts for the same LLM call budget.} EvoEngineer demonstrates faster initial convergence, reaching positive rewards within the first few generations while EoH struggles to escape near-zero performance early on. Both methods show continued improvement throughout the evolution process, indicating that LLM-driven search can progressively refine policy quality. The shaded regions reveal high variance across seeds, reflecting the stochastic nature of LLM-based code generation.
Different random seeds lead to different evolutionary trajectories, though EvoEngineer consistently outperforms EoH across all runs.

\subsection{Sample Efficiency Analysis}

To compare LLM-driven evolution with RL under equivalent environment interaction budgets, we conducted an experiment where EvoEngineer runs for 200 LLM calls (50 generations), consuming approximately 1M environment steps to match PPO's budget. Table~\ref{tab:sample_efficiency} summarizes this comparison.

\begin{table}[t]
    \centering
    \caption{Comparison at equivalent environment interactions ($\sim$1M steps).}
    \label{tab:sample_efficiency}
    \begin{tabular}{lccc}
        \hline
        \textbf{Method} & \textbf{Env Steps} & \textbf{Reward} & \textbf{Success\%} \\
        \hline
        PPO             & 1M                 & 214             & 60\%               \\
        EvoEngineer+    & $\sim$1M           & 144             & 70\%               \\
        \hline
    \end{tabular}
\end{table}

At equivalent sample budgets, EvoEngineer+ achieves a 70\% success rate compared to PPO's 60\%, though PPO attains higher average reward (214 vs 144) due to more consistent episode performance. This trade-off reflects a fundamental difference: PPO optimizes for expected return through gradient descent, while code evolution explores discrete policy structures that may excel in some episodes but fail in others.

However, this environment-step comparison has important limitations.
In simulated environments, environment interactions are computationally cheap; the true cost of LLM-driven evolution lies in API calls to large language models, not in running episodes.
With 200 LLM calls at current pricing, EvoEngineer+ incurs non-trivial inference costs, whereas
PPO's 1M environment steps require only local GPU computation.
This distinction becomes less significant as LLM inference costs continue to decrease, and reverses entirely in real-world robotics where physical environment interactions are expensive and potentially dangerous.

\subsection{Evolved Policy Analysis}

We analyze the best evolved policy (from EvoEngineer+, reward 143.6, 70\% success rate) to understand its control strategy. The policy implements a three-phase altitude-based approach, as shown in Figure~\ref{fig:policy_code}. The evolved code exhibits clear interpretable patterns: phase-based control with different logic for high, mid, and low altitudes; explicit human-readable thresholds for state conditions; and direct mapping from state observations to discrete actions. Unlike neural network weights, every decision boundary can be inspected and modified by human engineers.

\begin{figure}[t]
    \begin{lstlisting}
def policy(state):
  x, y, vx, vy, angle, w, left_leg, right_leg = state

  # Phase 1: High altitude (y > 0.6) - control descent
  if y > 0.6:
    if vy < -1.0: return 2           # main engine: brake
    if angle > 0.05 or w > 0.1:
      return 3                       # right engine: correct tilt
    if angle < -0.05 or w < -0.1:
      return 1                       # left engine: correct tilt

  # Phase 2: Mid altitude (0.2 < y <= 0.6) - correct drift
  elif y > 0.2:
    if vy < -0.5: return 2           # slow descent
    if x > 0.1 and vx > 0.15:
      return 3                       # correct rightward drift
    if x < -0.1 and vx < -0.15:
      return 1                       # correct leftward drift
    if angle > 0.05: return 3        # attitude correction
    if angle < -0.05: return 1

  # Phase 3: Low altitude (y <= 0.2) - soft landing
  else:
    if vy < -0.2: return 2           # ensure soft touchdown
    if x > 0.1: return 3             # fine position adjustment
    if x < -0.1: return 1

  return 0  # do nothing: conserve fuel
\end{lstlisting}
    \caption{Best evolved policy (simplified excerpt). The full policy contains additional edge-case handling for angular velocity and leg contact states.}
    \label{fig:policy_code}
\end{figure}




%% file: secs/06-discussion.tex
\section{Discussion}

\subsection{The Role of Context in LLM-Driven Evolution}

Our results highlight the critical importance of providing rich, structured context to guide LLM-based code evolution. FunSearch and EoH, despite using sophisticated evolutionary operators, struggle to discover effective control strategies because they lack domain-specific feedback. The LLM generates syntactically valid code but cannot reason about \textit{why} a policy fails or \textit{how} to improve it without explicit signals.

EvoEngineer addresses this gap by feeding episode statistics, failure mode analysis, and domain hints back to the LLM. When the model observes that crashes correlate with high descent velocity at low altitude, it can hypothesize that earlier braking is needed and modify thresholds accordingly. This closed-loop feedback transforms the LLM from a blind code mutator into an informed policy designer. The 7$\times$ performance gap between EvoEngineer and simpler approaches (66.6 vs. $\sim$8 average reward) underscores that context, not just evolutionary structure, drives effective policy synthesis.

\subsection{When to Choose Evolved Code over Neural Networks}

The comparison between EvoEngineer+ and PPO reveals a nuanced trade-off. PPO achieves higher average reward (214 vs. 144) because gradient-based optimization produces smooth policies that gracefully handle edge cases. However, EvoEngineer+ achieves a higher success rate (70\% vs. 60\%) because discrete decision boundaries can execute precise maneuvers when conditions are met, even if they fail abruptly otherwise.

This distinction matters in practice. For applications where task completion is binary (landed or crashed, package delivered or lost, patient stabilized or not), success rate may be the more relevant metric. For applications requiring consistent performance across all episodes, average reward better captures policy quality. Evolved code policies are particularly attractive when human oversight is required: a 59-line Python function can be reviewed, audited, and certified in ways that a 10K-parameter neural network cannot. Domain experts can directly inspect decision boundaries, verify safety constraints, and modify thresholds without retraining.

\subsection{Scaling and Practical Considerations}

Scaling from 45 to 200 LLM calls improves reward from 66.6 to 143.6 and success rate from 40\% to 70\%, while simultaneously reducing code length from 80 to 59 lines. This pattern suggests that early iterations capture coarse-grained improvements, while later iterations refine edge cases and eliminate redundant logic. Notably, extended evolution produces not only better-performing policies but also simpler ones, indicating that the LLM learns to distill control knowledge into more elegant representations.

The cost structure differs fundamentally from neural network training. LLM API calls dominate the budget for code evolution, while environment interactions are essentially free. For PPO, the situation reverses: GPU computation is the primary cost, and simulation steps are cheap. This makes code evolution particularly attractive for domains where environment interaction is expensive or dangerous, such as physical robotics, where each real-world episode carries time, wear, and safety costs that dwarf API expenses.

\subsection{Limitations}

Several limitations constrain the current work. Our evaluation focuses on a single benchmark (LunarLander-v3) with discrete actions; generalization to continuous control, higher-dimensional state spaces, and more complex tasks remains to be demonstrated. The approach inherits the LLM's biases and capabilities, meaning performance may vary across models and could degrade for domains outside the LLM's training distribution. Evolved policies exhibit higher variance than neural networks, succeeding decisively or failing completely rather than achieving consistent moderate performance. Finally, while LLM inference costs continue to decrease, extended evolution (200+ calls) still incurs non-trivial API expenses compared to local GPU training.

\section{Conclusion}

We presented a systematic study of LLM-driven code evolution for control policy synthesis. By comparing three evolution strategies on the LunarLander benchmark, we demonstrated that rich contextual feedback (EvoEngineer) dramatically outperforms minimal-prompt approaches (FunSearch, EoH), achieving 7$\times$ higher reward with the same LLM budget. Extended evolution (EvoEngineer+) produces interpretable 59-line policies that achieve 70\% success rate, exceeding PPO's 60\% despite lower average reward. The key insight is that LLMs can synthesize effective control logic when provided with structured feedback about behavior, not just fitness scores.

Future work includes extending to continuous action spaces, evaluating on complex locomotion and manipulation tasks, exploring hybrid approaches that combine LLM evolution with gradient-based fine-tuning, and investigating multi-objective evolution that jointly optimizes performance, interpretability, and robustness.